\documentclass[10pt,journal,compsoc]{IEEEtran}
\IEEEoverridecommandlockouts

\usepackage[utf8]{inputenc}
\usepackage[T1]{fontenc}
\usepackage{microtype}
\usepackage{graphicx}
\usepackage{booktabs}
\usepackage{multirow}
\usepackage{siunitx}
\usepackage{makecell}
\usepackage{mathtools}
\usepackage{tabularx,array}
\usepackage[table,dvipsnames,svgnames,x11names]{xcolor}
\usepackage{wrapfig}
\usepackage{xspace}
\usepackage{amsmath,amssymb,amsfonts}
\usepackage{amsthm}
\usepackage{algorithm}
\usepackage{algpseudocode}
\usepackage{epsfig}
\usepackage{etoolbox}
\usepackage{listings}
\usepackage{enumitem}
\usepackage{graphbox}
\usepackage[breakable]{tcolorbox}
\usepackage{url}
\usepackage{caption}
\usepackage{placeins}
\usepackage{bm}
\usepackage{cite}
\usepackage[pdftex,breaklinks=true,bookmarks=true]{hyperref}
\usepackage[capitalise,nameinlink,noabbrev]{cleveref}

\definecolor{projectpurple}{RGB}{231, 70, 151}

\hypersetup{
    colorlinks=true,
    linkcolor=blue,
    anchorcolor=blue,
    citecolor=blue,
    urlcolor=projectpurple,
    pdfborder={0 0 0}
}

\crefname{prompt}{Prompt}{Prompts}
\Crefname{prompt}{Prompt}{Prompts}

\newcommand{\myparagraph}[1]{\vspace{0.5em}\noindent\textbf{#1}}

\definecolor{lightblue}{rgb}{0.93, 0.96, 1.0}
\newcounter{prompt}
\renewcommand{\theprompt}{\arabic{prompt}}
\newcommand{\prompt}[3]{
\refstepcounter{prompt}
\begin{tcolorbox}[
    colback=lightblue!35,
    colframe=white!45!black,
    title={Prompt.~\theprompt:~#1},
    breakable,
]
#2
\label{#3}
\end{tcolorbox}
}

\begin{document}

\title{ThinkV2V: Unleashing the Reasoning Capability of MLLMs for Instruction-Guided Video Editing}

\author{Donghao Zhou*, Haoyang He*, Fan Zhang, Hao Yang\textsuperscript{\textdagger}, Guisheng Liu, Xin Gao, Zhongwei Wan,\\ Xingyuan Bu, Jie Wang, Qiangpeng Yang, Shilei Wen\textsuperscript{\S}, Chi-Wing Fu, and Pheng-Ann Heng\textsuperscript{\S}
\IEEEcompsocitemizethanks{
\IEEEcompsocthanksitem *Equal contributions. \textsuperscript{\textdagger}Project lead. \textsuperscript{\S}Corresponding authors.
\IEEEcompsocthanksitem Donghao Zhou, Fan Zhang, Chi-Wing Fu, and Pheng-Ann Heng are with The Chinese University of Hong Kong.
\IEEEcompsocthanksitem Haoyang He is with Zhejiang University.
\IEEEcompsocthanksitem Hao Yang, Guisheng Liu, Xin Gao, Xingyuan Bu, Jie Wang, Qiangpeng Yang, and Shilei Wen are with ByteDance.
\IEEEcompsocthanksitem Zhongwei Wan is with The Ohio State University.
}}

\IEEEtitleabstractindextext{
\begin{abstract}

Instruction-guided video editing has made significant progress, yet existing methods use multimodal large language models (MLLMs) primarily as semantic encoders, so they often fall short in working with implicit edits that require causal or semantic reasoning.
To bridge this fundamental gap in video editing, we propose \textbf{ThinkV2V}, a reasoning-driven framework for complex instruction-guided video editing, explicitly activating MLLM thinking before visual generation.
At its core, ThinkV2V builds on a practical MLLM-to-DiT architecture to turn explicit thinking over the source video and instruction into refined conditioning signals for video editing.
Further, we equip it with a dedicated training and inference recipe, combining \textit{Progressive Curriculum Training}, which gradually cultivates the model from basic editing to reasoning-intensive cases, with \textit{Inference-Time Thinking Scaling}, which iteratively refines candidate prompts and selects the most reliable one, to better elicit reasoning in challenging editing scenarios.
We also curate the \textit{ThinkV2V-150K} dataset and introduce \textit{ThinkV2V-Bench} to support training and evaluation of video editing with implicit intent and causal reasoning.
Experimental results demonstrate the state-of-the-art performance of ThinkV2V on both complex and standard editing scenarios, in which our 5B-scale DiT model substantially outperforms larger 10B-scale baselines.
The project page is available at \url{https://correr-zhou.github.io/ThinkV2V}.

\end{abstract}

}

\maketitle

\section{Introduction}

\begin{figure*}[t]
    \centering
    
    \includegraphics[width=1\linewidth]{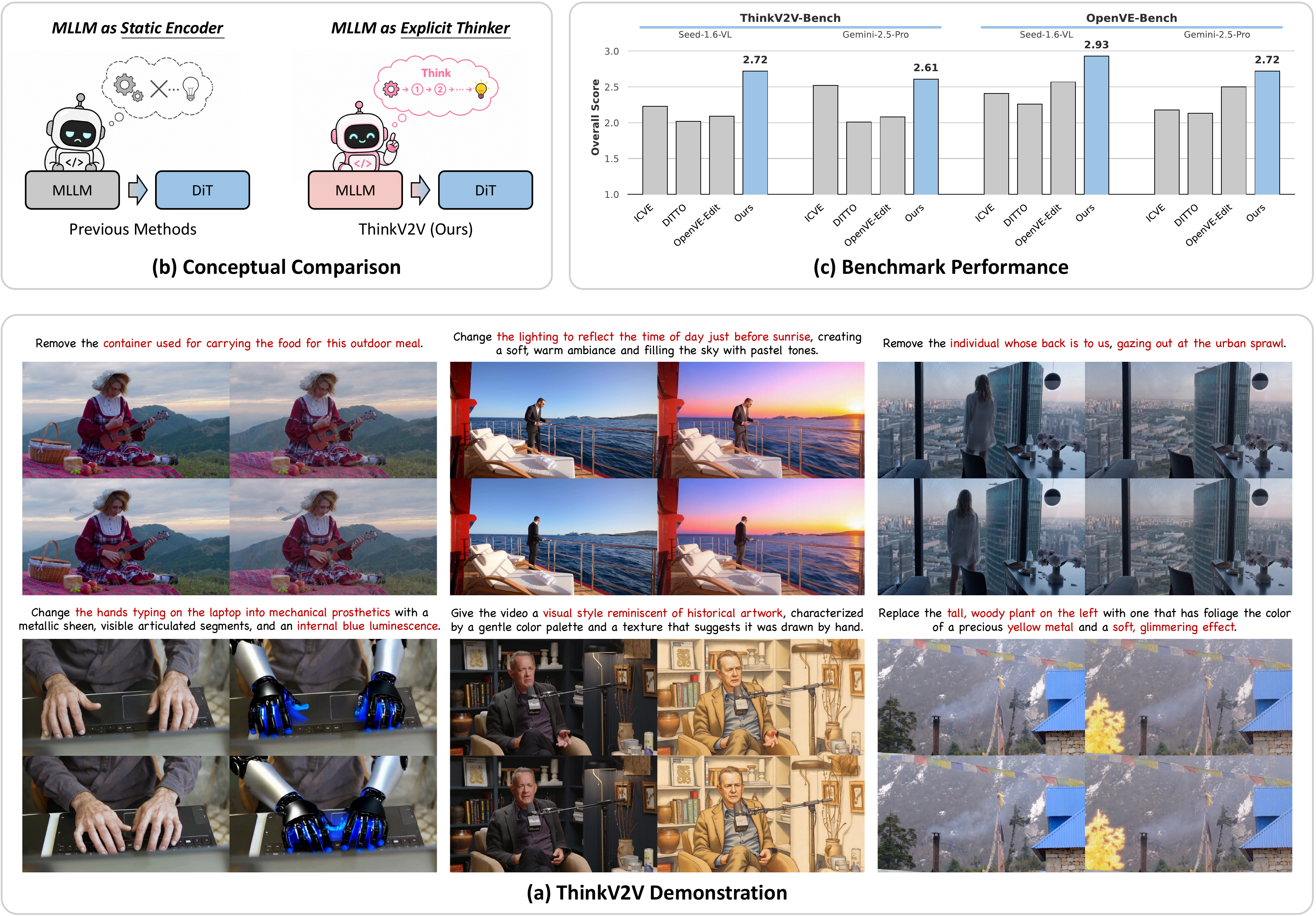}

    \caption
    {
        \textbf{Overview of ThinkV2V.}
        (a) Documentation on video editing with complex instructions, illustrating the capabilities of ThinkV2V.
        (b) Conceptual comparison between prior methods that mainly use MLLMs as semantic encoders and our method that activates the MLLM as an explicit thinker.
        (c) Quantitative results showing the strong overall performance of ThinkV2V on complex and standard editing scenarios.
        \textit{Zoom in for better view.}
    }

    \label{fig:teaser}

\end{figure*}

\IEEEPARstart{R}{ecent} advances in diffusion models and multimodal large language models (MLLMs) have greatly expanded instruction-guided visual generation across diverse scenarios, enabling users to manipulate images and videos with natural language rather than handcrafted control signals~\cite{kontext, zhou2024magictailor, omnigen2, liu2026hifi, qwenimage, openve, zhou2026identitystory, omnivideo, zhou2026omnishow, univideo}.
In the image domain, recent systems have shown that visual generation moves toward flexible instruction following and intent understanding~\cite{kontext, song2025scenedecorator, omnigen2, chen2025empirical, qwenimage}.
As this trend extends to video, the scope is no longer merely text-conditioned generation, but chatting-style editing that can more faithfully capture user intent~\cite{openve, omnivideo, univideo, icve, lucyedit}.
However, many user-provided video editing instructions are not direct descriptions of a target appearance, while they are often implicit, indirect, and require causal or semantic reasoning to reveal the actionable editing goals (\cref{fig:teaser}(a)).
Consequently, real-world video editing is not merely a generation problem, but a cognition-intensive process that requires instruction understanding, reasoning, and execution to work in concert.

These more realistic editing demands expose the fundamental limitations of current video editing paradigms.
Although recent instruction-guided video editors have achieved encouraging progress on simple, directly specified edits~\cite{openve, univideo, omnivideo}, most existing methods still operate as ``black-box'' pipelines.
In particular, \textit{they primarily use MLLMs as stronger semantic encoders for jointly processing the prompt and the source video, rather than fully unleashing their explicit thinking or reasoning capabilities} (\cref{fig:teaser}(b)).
As a result, they remain effective for explicit prompts, yet often struggle with instructions whose true intent is implicit and can only be grounded correctly through causal reasoning.
In such cases, the model may latch onto surface keywords while missing the actual editing objective, especially when the instruction involves temporal dynamics, object relations, or logical causality.
Therefore, the bottleneck of current methods is not merely insufficient generation quality, but the lack of an explicit ``think-before-edit'' process that can transform complex language into a reliable editing scheme.

However, moving from ``perception-driven'' editing to ``reasoning-driven'' editing naturally raises three key challenges.
First, explicit thinking generated by an MLLM contains high-level semantic and causal information, but preserving these signals and stably translating them into conditions that genuinely benefit video generation is difficult to achieve within existing editing frameworks.
Second, complex video editing naturally incurs higher comprehension difficulty and a higher retry cost, which means that standard training alone or a single inference pass is insufficient to fully unlock the reasoning potential of MLLMs.
Third, current datasets and benchmarks mainly emphasize basic editing ability, leaving a clear gap in training and evaluating video editors on instructions that require implicit intent understanding, causal reasoning, and other high-density reasoning capabilities.
To address these issues, we present \textbf{ThinkV2V}, a reasoning-driven framework for complex instruction-guided video editing that explicitly activates MLLM thinking before visual generation (\cref{fig:teaser}(b)).

First, we design a practical \textit{MLLM-to-DiT architecture} built upon Qwen3-VL-Thinking-8B~\cite{bai2025qwen3}, a connector with learnable query tokens, and Wan-2.1~\cite{wan2025wan}, where the MLLM takes the source video and the original instruction as input, produces a refined prompt through explicit thinking, and provides hidden features that are further refined by the connector and injected into the DiT.
To avoid an overly narrow information bottleneck between the MLLM and the DiT, ThinkV2V further retains sufficient visual and textual context by feeding both the source-video VAE features and the original prompt embeddings into the DiT.

Second, beyond the architecture itself, we further introduce a dedicated \textit{training-and-inference recipe} to better unlock MLLM reasoning for instruction-guided video editing.
Specifically, it combines \textit{Progressive Curriculum Training}, which gradually adapts the model from stable basic editing toward reasoning-intensive editing, with \textit{Inference-Time Thinking Scaling}, which iteratively refines candidate prompts and lets the MLLM itself select the most reliable one before the DiT process, together offering a more effective mechanism for eliciting reasoning in complex editing scenarios.

Finally, to support this reasoning-driven paradigm, we curate the \textit{ThinkV2V-150K} dataset and introduce \textit{ThinkV2V-Bench}, providing dedicated data and evaluation for complex video editing with implicit intent and causal reasoning.
Extensive experiments show that ThinkV2V establishes state-of-the-art performance on complex and standard editing scenarios across diverse editing categories (\cref{fig:teaser}(c)), with our 5B-scale DiT model even surpassing several substantially larger 10B-scale baselines.

Our main contributions are summarized as follows:
\begin{itemize}[itemsep=0.38em, topsep=0.2em]
    \item We propose \textbf{ThinkV2V}, a reasoning-driven video editing framework with a practical \textit{MLLM-to-DiT architecture} that explicitly incorporates MLLM thinking for faithful understanding and planning before generation.
    \item We introduce a dedicated recipe for reasoning-enhanced video editing, consisting of \textit{Progressive Curriculum Training} and \textit{Inference-Time Thinking Scaling}, which continuously improves how reasoning is learned during training and exploited at inference time.
    \item We curate \textit{ThinkV2V-150K} and present \textit{ThinkV2V-Bench}, providing a systematic training and evaluation foundation for complex video editing centered on implicit intent understanding and causal reasoning.
    \item We demonstrate through extensive experiments that ThinkV2V outperforms existing methods on reasoning-driven video editing scenarios while remaining competitive on standard editing tasks.
\end{itemize}

\section{Related Work}
\myparagraph{Instruction-guided Image and Video Editing.}
Advances in diffusion models have propelled research in image and video editing. Instruction-guided image editors are largely data-driven. Leveraging pre-trained priors, they typically concatenate VAE or SigLIP encoded features with noise latents along either the sequence dimension, as in ImgEdit~\cite{imgedit}, Step1X-Edit~\cite{step1xedit}, FLUXKontext~\cite{kontext}, OmniGen2~\cite{omnigen2}, and Qwen-Image-Edit~\cite{qwenimage}, or the channel dimension in X2Edit~\cite{x2edit}, followed by large-scale training.
Instruction-guided video editing has surged, exploring diverse feature integration strategies. These include sequence concatenation in Omni-Video~\cite{omnivideo}, UniVideo~\cite{univideo}, and ICVE~\cite{icve}, channel concatenation in Lucy-Edit~\cite{lucyedit} and OpenVE~\cite{openve}, and element-wise addition in InstructX~\cite{instructx}. To enhance flexibility, Kiwi-Edit~\cite{kiwiedit} introduces dual Instruction and Reference Guidance. Beyond feature fusion, SAMA~\cite{sama} decouples the editing process into Semantic Anchoring and Motion Alignment to mitigate feature interference. Overall, these methods mainly improve how semantic features are injected into video generators, but still leave limited support for explicitly reasoning over implicit user intent before editing.

\myparagraph{Reasoning-Enhanced Visual Generation and Editing.}
Multimodal large language models (MLLMs) have shifted from basic perception to System-2 visual reasoning. To address limitations of traditional models in handling complex spatial relationships and logical causality, recent works introduce explicit cognitive mechanisms. LLaVA-CoT~\cite{llavacot} proposes a framework to generate structured Chains-of-Visual-Thought, while BLINK-Twice~\cite{ye2025blink} highlights that fine-grained analytical reasoning is indispensable for multimodal understanding. In visual generation, integrating reasoning into diffusion models is crucial for complex scene synthesis. For image generation, RPG~\cite{yang2024mastering} leverages LLM chain-of-thought capabilities for multi-region layout planning. In video generation, where physical consistency matters, VChain~\cite{huang2025vchain} injects MLLM reasoning signals during inference to construct a visual chain-of-thought guiding frame evolution. Similarly, Hao \textit{et al.}~\cite{hao2025enhancing} utilize counterfactual reasoning to evaluate implausibility, guiding generation away from physics-violating trajectories. 
Despite these advances, the systematic extension of reasoning to video editing remains underexplored. Existing instruction-guided methods are confined to perception-level manipulations (\textit{e.g.}, style transfer) and fail on tasks requiring spatial-temporal reasoning. Although prior work has taken an initial step toward self-reflective editing~\cite{revise}, it lacks a mechanism to decouple the MLLM's explicit cognitive process from the underlying video generator.

\section{Methodology}

We present \textbf{ThinkV2V}, a video editing framework that explicitly activates MLLM thinking over the source video and instruction, and then effectively turns the resulting features into conditioning signals for video editing.
We start by introducing our \textit{MLLM-to-DiT architecture} that comprises an MLLM, a learnable-query connector, and a multi-condition DiT (\cref{sec:method:arch}).
Then, we describe our \textit{training-and-inference recipe} featuring \textit{Progressive Curriculum Training} and \textit{Inference-Time Thinking Scaling} (\cref{sec:method:recipe}).
Finally, we present reasoning-oriented dataset and benchmark for instruction-guided video editing, including \textit{ThinkV2V-150K} and \textit{ThinkV2V-Bench} (\cref{sec:method:data}).

\begin{figure*}[t]
    \centering
    
    \includegraphics[width=1\linewidth]{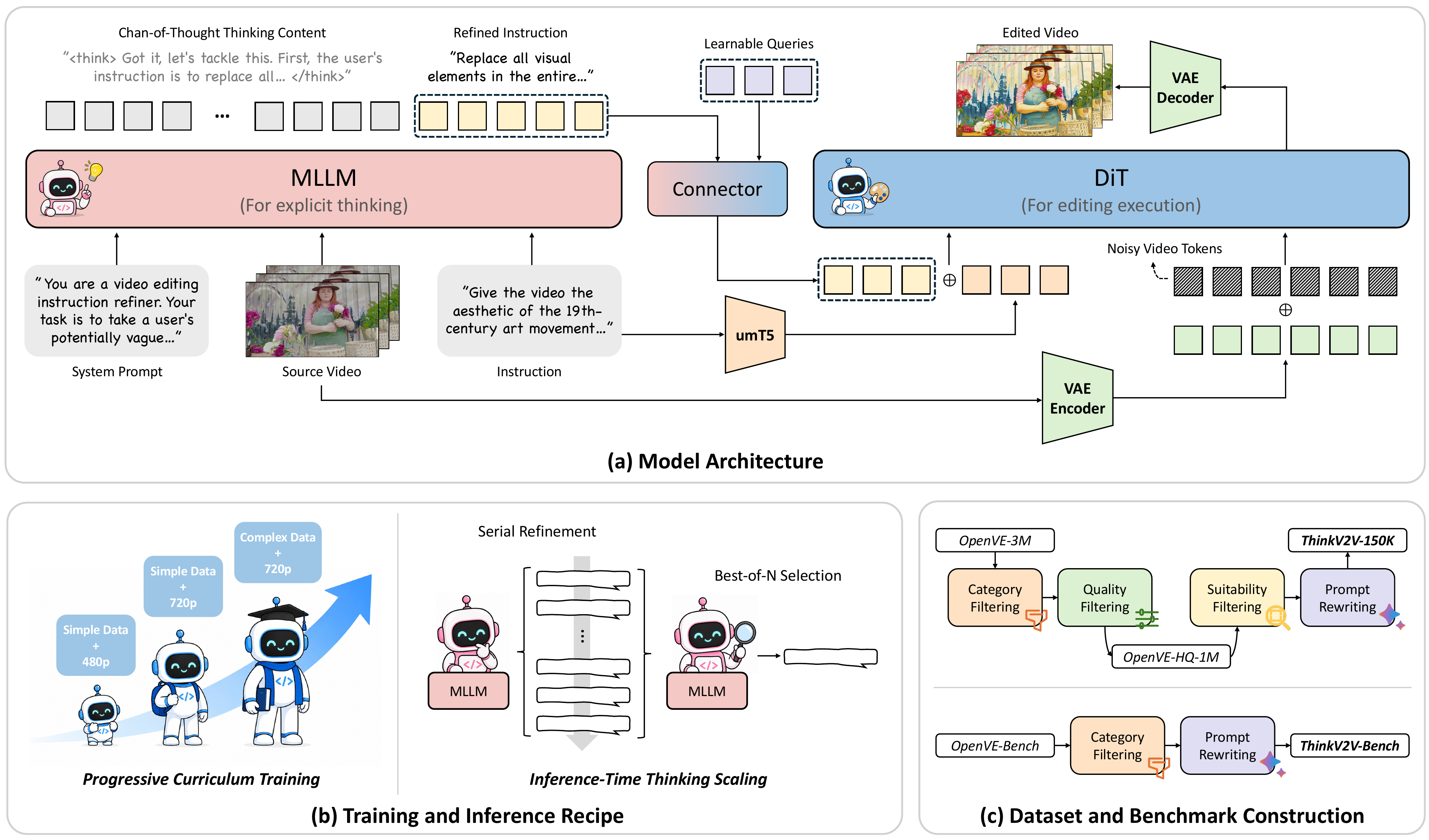}

    \caption
    { 
        \textbf{Pipeline of ThinkV2V.}
The ThinkV2V framework consists of three key parts:
\textit{(a) Model architecture} (\cref{sec:method:arch}) uses an MLLM for explicit thinking, a learnable-query connector for feature alignment, and a multi-condition DiT for final video editing.
\textit{(b) Training and inference recipe} (\cref{sec:method:recipe}) includes Progressive Curriculum Training and Inference-Time Thinking Scaling to better elicit reasoning for complex editing.
\textit{(c) Dataset and benchmark construction} (\cref{sec:method:data}) for ThinkV2V-150K and ThinkV2V-Bench aims to support reasoning-oriented training and evaluation.
    }
    
    \label{fig:pipeline}
    
\end{figure*}

\subsection{Model Architecture}
\label{sec:method:arch}

\myparagraph{MLLM for Explicit Thinking.}
In ThinkV2V, the MLLM takes the source video and the original instruction as input, and performs Chain-of-Thought (CoT) reasoning to model the implicit intent and causal relations behind complex editing requests.
We adopt Qwen3-VL-Thinking-8B~\cite{bai2025qwen3} as the MLLM, and use a dedicated system prompt to encourage it to first produce intermediate thinking and then output a semantically precise refined prompt that is expected to be more suitable for subsequent editing.
Formally, this thinking-and-refinement process is written as
\begin{equation}
    (\mathcal{T}, \hat{\mathcal{I}}, \mathbf{H}_{\mathrm{ans}})
    =
    f_{\mathrm{MLLM}}(\mathbf{V}, \mathcal{I}; \mathcal{P}_{\mathrm{sys}}),
    \label{eq:mllm_thinking}
\end{equation}
where $\mathbf{V}$ is the source video, $\mathcal{I}$ is the original instruction, $\mathcal{P}_{\mathrm{sys}}$ is the system prompt, $\mathcal{T}$ denotes the generated thinking content, $\hat{\mathcal{I}}$ is the refined prompt, and $\mathbf{H}_{\mathrm{ans}}$ denotes the hidden states of the answer tokens after the \texttt{</think>} tag. Note that only $\mathbf{H}_{\mathrm{ans}}$ is passed to the following modules.
This ``thinking-then-refinement'' process is important because many user-provided editing instructions are implied, and often become actionable only after causal reasoning.
In addition to the refined prompt text, we extract the last-layer hidden states corresponding to the tokens generated after the \texttt{</think>} tag, and treat them as high-level semantic features for video editing.
With this design, the MLLM serves as an explicit reasoner for real-world instruction understanding, rather than a static semantic encoder.

\myparagraph{Learnable-Query Connector.}
We place a connector between the MLLM and the DiT to translate MLLM semantic features into a fixed-length conditioning representation that can be consumed stably for video editing.
Concretely, the connector maintains a set of learnable query tokens, and uses cross-attention to extract a compact feature sequence from the MLLM hidden states.
Given the answer-token hidden states $\mathbf{H}_{\mathrm{ans}}$ from \cref{eq:mllm_thinking}, the connector produces
\begin{equation}
    \mathbf{C}
    =
    \operatorname{CrossAttn}
    \left(
    \mathbf{Q}_{\mathrm{learn}},
    \mathbf{K}_{\mathrm{ans}},
    \mathbf{V}_{\mathrm{ans}}
    \right)
    \in \mathbb{R}^{N_q \times d_c},
    \label{eq:query_connector}
\end{equation}
where $\mathbf{Q}_{\mathrm{learn}} \in \mathbb{R}^{N_q \times d_c}$ denotes the learnable queries, $\mathbf{K}_{\mathrm{ans}}$ and $\mathbf{V}_{\mathrm{ans}}$ are key-value projections from $\mathbf{H}_{\mathrm{ans}}$, $N_q$ is the query length, $d_c$ is the connector feature dimension, and $\mathbf{C}$ is the fixed-length conditioning feature sequence passed to the DiT.
This learnable-query mechanism performs both feature compression and cross-module alignment, mapping MLLM-derived representations into a DiT-friendly conditioning space.
Compared with directly exposing the DiT to a long and potentially unstable token sequence, learnable queries can actively aggregate the most useful information for editing and improve robustness of the feature interaction across modules.
The extracted fixed-length semantic features are then injected into the subsequent diffusion process as core high-level conditions.

\myparagraph{Multi-Condition DiT.}
We instantiate the DiT backbone with Wan-2.1-5B~\cite{wan2025wan}, which serves as the execution module that produces the final edited video under multiple complementary conditions.
First, the fixed-length features produced from the connector provide high-level editing intent derived from explicit thinking.
Second, we additionally feed the original-instruction text embeddings and the source-video VAE features into the DiT to preserve sufficient textual and visual context.
Specifically, the original-instruction text embeddings are concatenated with the connector features along the token dimension and injected into the DiT through cross-attention, while, inspired by LucyEdit, the source-video VAE features are fused with the noisy video latents through channel concatenation.
The multi-condition denoising step can be summarized as
\begin{equation}
    \hat{\mathbf{z}}_0
    =
    f_{\mathrm{DiT}}
    \left(
    t,
    \mathbf{z}_t \mathbin{\Vert_{\mathrm{ch}}} \mathbf{z}_{\mathrm{src}},
    \mathbf{C} \mathbin{\Vert_{\mathrm{tok}}} \mathbf{E}_{\mathrm{text}}
    \right),
    \label{eq:multi_condition_dit}
\end{equation}
where $t$ is the time-step embedding, $\mathbf{z}_t$ is the noisy video latent, $\mathbf{z}_{\mathrm{src}}$ is the source-video VAE latent, $\mathbf{E}_{\mathrm{text}}$ denotes text embeddings from the original instruction $\mathcal{I}$, and $\Vert_{\mathrm{tok}}$ and $\Vert_{\mathrm{ch}}$ denote token-wise and channel-wise concatenation, respectively.
Using $\mathcal{I}$ avoids the influence of simply using the refined prompt as text input and isolates the contribution of $\mathbf{C}$.
This multi-condition design mitigates the information bottleneck when bridging the MLLM to DiT, and avoids over-reliance on the compressed MLLM features when fine-grained content details are required.
As a result, the DiT acts as a multi-condition video editor that follows reasoning-level guidance while maintaining low-level visual fidelity.

\begin{figure*}[t]
    \centering
    
    \includegraphics[width=1\linewidth]{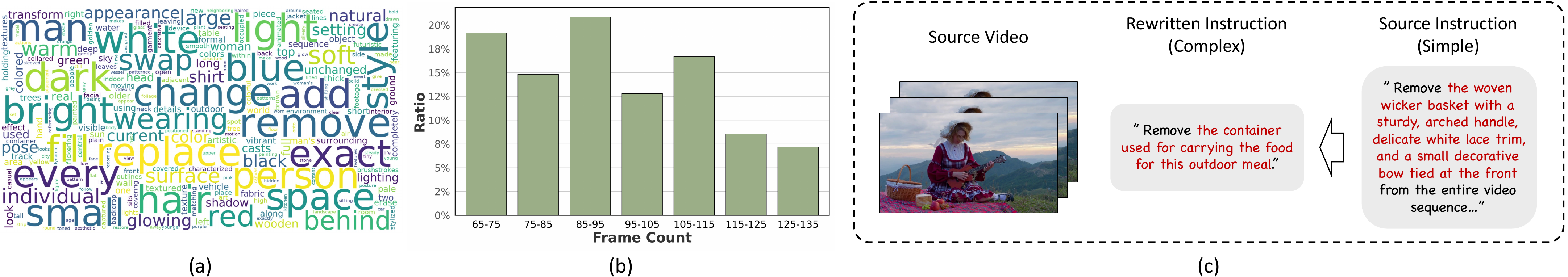}
    
    \caption
    { 
        \textbf{Statistics and example of ThinkV2V-150K and ThinkV2V-Bench.}
        (a) Word cloud of complex instructions, illustrating the semantic diversity of our reasoning-oriented editing data.
        (b) Distribution of source-video frame counts, showing the temporal diversity of the collected videos.
        (c) A representative benchmark sample, including the source video and the corresponding instructions.
    }
    
    \label{fig:data_bench}
    
\end{figure*}

\subsection{Training and Inference Recipe}
\label{sec:method:recipe}

\myparagraph{Progressive Curriculum Training.}
Beyond architecture, we need a training recipe that progressively cultivates the ability to translate high-level reasoning semantics into stable video editing behavior.
Video editing could become harder as both resolution and instruction complexity increase, and directly training on high-resolution complex instructions can make it difficult to learn spatial-temporal alignment and reasoning-driven execution simultaneously.
We therefore adopt \textit{Progressive Curriculum Training} that progresses along two axes, from low to high resolution and from simple to complex instructions, and progressively organizes training into three stages:
\begin{enumerate}[label={(\arabic*)}, leftmargin=2em, itemsep=-0.2em, topsep=-0.2em]
    \item \textit{Stage~1 (Basic Alignment)} focuses on learning a stable mapping from MLLM features to the DiT conditioning space, so that the model can reliably perform basic edits.
    \item \textit{Stage~2 (High-Resolution Adaptation)} focuses on scaling up resolution and consolidating high-resolution editing quality, improving visual fidelity and stability.
    \item \textit{Stage~3 (Reasoning-Intensive Tuning)} continues training on complex instructions to better handle implicit intent, causal relations, and other high-density reasoning requirements.
\end{enumerate}
In practice, the curriculum moves from the simple-instruction dataset OpenVE-HQ-1M (for Stage 1\&2) to the complex-instruction dataset ThinkV2V-150K (for Stage 3), with data details described in \cref{sec:method:data}.
During this stage-wise training, we primarily optimize the connector and the DiT, while keeping the MLLM frozen to preserve its general thinking capability and to avoid unnecessary interference with multimodal understanding.

\myparagraph{Inference-Time Thinking Scaling.}
After meticulous training, there remains headroom to further elicit MLLM reasoning at test time, which can further improve the final editing performance on challenging instructions.
For implicit or causality-heavy requests, a single prompt refinement may be suboptimal because the reasoning can miss key constraints or drift toward an incorrect interpretation.
To address this, we introduce \textit{Inference-Time Thinking Scaling}, which first performs serial refinement by feeding the refined prompt back into the MLLM for multiple rounds to produce a sequence of evolving candidates.
Since serial refinement alone may amplify early mistakes and yield final results that are internally consistent but off-target, we further apply a best-of-N selection step ($N=8$), where the MLLM chooses the candidate that best matches the original editing intent and source video.
We then use the cached last-layer hidden states of the selected candidate and pass them to the connector and DiT to generate the edited video.
Empirically, this post-hoc strategy often brings stable gains even without additional explicit training, which also reflects the advantage of treating the MLLM as a reasoner rather than just an encoder, since its explicitly generated texts can be naturally reused.

\subsection{Dataset and Benchmark Construction}
\label{sec:method:data}

\myparagraph{Dataset: ThinkV2V-150K.}
To train reasoning-driven video editors under realistic user-intent distributions, we require data that captures implicit goals and causal relations beyond explicit target appearances.
We construct ThinkV2V-150K from OpenVE-3M, retaining only five categories: ``Global Style'', ``Background Change'', ``Local Remove'', ``Local Add'', and ``Local Change''.
We exclude the other three categories, such as ``Subtitle Edit'', as they are less suitable for reasoning-centric editing.
We first filter 2M samples from the selected five categories, rescore them with Gemini-2.5-Flash, and compute the average inter-frame CLIP similarity (CLIP-F)~\cite{radford2021learning} and Temporal Flickering (TF)~\cite{huang2024vbench} for each edited video.
We retain the top 50\% highest-quality samples to form OpenVE-HQ-1M for Stage 1 and Stage 2 training.
From OpenVE-HQ-1M, we further select 20K to 40K samples per category that are suitable for reasoning-instruction synthesis, and use Gemini-2.5-Pro~\cite{comanici2025gemini} to generate complex reasoning-oriented instructions that approximate real-world implicit user requests for the resulting 150K video pairs.
ThinkV2V-150K thus exposes models to instruction distributions closer to real user requests and promotes editing grounded in implicit intent and causal reasoning.

\myparagraph{Benchmark: ThinkV2V-Bench.}
Standard benchmarks mainly assess basic editing quality and are insufficient for testing whether a model follows implicit intent.
We therefore construct ThinkV2V-Bench to evaluate the translation of complex language understanding into executable editing conditions.
Consistent with ThinkV2V-150K, we benchmark only the same five reasoning-suitable categories.
For each source video, edited video, and instruction in OpenVE-Bench, we use Gemini-3.1-Pro~\cite{geminiteam2026gemini31pro} to rewrite the original direct instruction into a reasoning-oriented one.
This process yields 308 video-instruction pairs across five categories, forming a compact benchmark for validating reasoning-driven video editing.
In \cref{fig:data_bench}, we summarize the statistics of our dataset and benchmark, and also present a representative benchmark example.
Together, ThinkV2V-150K and ThinkV2V-Bench form a data-and-evaluation loop that enables systematic training and validation of our core claim on reasoning-driven video editing.

\section{Experiments}

\begin{table*}[t]
    \centering
    \caption{
     \textbf{Quantitative comparison.}
     We conduct quantitative evaluation on both ThinkV2V-Bench and OpenVE-Bench~\cite{he2025openve}, using two judge models, Seed-1.6-VL~\cite{seed2025introduction} and Gemini-2.5-Pro~\cite{comanici2025gemini}. ``\#Param.'' denotes the parameter count of the video editor. The results show that our method achieves competitive overall performance across both benchmarks and all evaluated settings, indicating its consistent effectiveness across diverse editing scenarios.
     }

    \setlength{\tabcolsep}{2.5mm}
    \renewcommand{\arraystretch}{1.2}
    
    \resizebox{\linewidth}{!}
    {
\newcommand{\methodcolwidth}{3cm}
\newcommand{\scorecolwidth}{2.4cm}
\newcommand{\headerrowheight}{2.8ex}
\begin{tabular}{m{\methodcolwidth}|c|*{5}{>{\centering\arraybackslash}m{\scorecolwidth}}|c}
\toprule
\rule{0pt}{\headerrowheight}\textbf{Method} & \textbf{\#Param.} & \textbf{Global Style $\uparrow$} & \textbf{Bg. Change $\uparrow$} & \textbf{Local Change $\uparrow$} & \textbf{Local Remove $\uparrow$} & \textbf{Local Add $\uparrow$} & \textbf{Overall $\uparrow$} \\
\midrule
\multicolumn{8}{c}{{ThinkV2V-Bench}} \\
\midrule
\multicolumn{8}{c}{\textit{Evaluation on Seed-1.6-VL}} \\
VACE~\cite{vace}  & 14B   & 1.33  & 1.17  & 1.41  & 1.00  & 1.05  & 1.19 \\
OmniVideo~\cite{omnivideo} & 1.3B  & 1.92  & 1.40  & 1.91  & 1.96  & 1.87  & 1.82 \\
ReCo~\cite{zhang2025region} & 1.3B  & 2.25  & 1.62  & 2.50  & 3.13  & 2.39  & 2.26 \\
InsVIE~\cite{insvie} & 2B    & 2.08  & 1.18  & 1.48  & 1.04  & 1.24  & 1.40 \\
Lucy-Edit~\cite{lucyedit} & 5B    & 1.69  & 2.08  & 2.84  & 1.09  & 2.21  & 2.01 \\
ICVE~\cite{icve}  & 13B   & 1.95  & 2.04  & 2.91  & 2.05  & 2.12  & 2.23 \\
DITTO~\cite{ditto} & 14B   & 3.28  & 1.93  & 2.12  & 1.00  & 1.82  & 2.02 \\
OpenVE-Edit~\cite{he2025openve} & 5B    & 2.24  & 2.49  & 2.55  & 1.14  & 2.05  & 2.09 \\
\rowcolor[RGB]{227,240,251}
ThinkV2V (Ours) & 5B    & 3.13  & 2.67  & 3.07  & 2.30  & 2.42  & \textbf{2.72} \\
\midrule
\multicolumn{8}{c}{\textit{Evaluation on Gemini-2.5-Pro}} \\
VACE~\cite{vace}  & 14B   & 1.77  & 1.91  & 1.95  & 2.07  & 1.48  & 1.83 \\
OmniVideo~\cite{omnivideo} & 1.3B  & 2.48  & 1.18  & 1.25  & 1.56  & 1.23  & 1.52 \\
ReCo~\cite{zhang2025region} & 1.3B  & 2.58  & 1.60  & 2.03  & 2.71  & 2.42  & 2.27 \\
InsVIE~\cite{insvie} & 2B    & 2.90  & 1.18  & 1.46  & 1.38  & 1.29  & 1.63 \\
Lucy-Edit~\cite{lucyedit} & 5B    & 2.84  & 2.01  & 3.12  & 1.87  & 2.39  & 2.46 \\
ICVE~\cite{icve} & 13B   & 2.90  & 1.99  & 2.98  & 2.23  & 2.36  & 2.52 \\
DITTO~\cite{ditto} & 14B   & 3.83  & 1.36  & 1.97  & 1.39  & 1.57  & 2.01 \\
OpenVE-Edit~\cite{he2025openve} & 5B    & 3.06  & 2.05  & 2.06  & 1.32  & 1.90  & 2.08 \\
\rowcolor[RGB]{227,240,251}
ThinkV2V (Ours) & 5B    & 3.74  & 2.39  & 2.89  & 2.03  & 2.02  & \textbf{2.61} \\
\midrule
\multicolumn{8}{c}{{OpenVE-Bench}} \\
\midrule
\multicolumn{8}{c}{\textit{Evaluation on Seed-1.6-VL}} \\
VACE~\cite{vace}  & 14B   & 1.41  & 1.16  & 1.43  & 1.00  & 1.05  & 1.21 \\
OmniVideo~\cite{omnivideo} & 1.3B  & 1.02  & 1.00  & 1.00  & 1.00  & 1.00  & 1.00 \\
ReCo~\cite{zhang2025region} & 1.3B  & 2.68  & 1.69  & 2.67  & 2.46  & 2.42  & 2.38 \\
InsVIE~\cite{insvie} & 2B    & 2.25  & 1.23  & 1.60  & 1.00  & 1.23  & 1.46 \\
Lucy-Edit~\cite{lucyedit} & 5B    & 2.17  & 2.20  & 3.30  & 1.03  & 2.37  & 2.21 \\
ICVE~\cite{icve}  & 13B   & 2.35  & 1.86  & 2.91  & 2.68  & 2.27  & 2.41 \\
DITTO~\cite{ditto} & 14B   & 3.70  & 2.23  & 2.28  & 1.00  & 2.08  & 2.26 \\
OpenVE-Edit~\cite{he2025openve} & 5B    & 3.11  & 2.72  & 3.19  & 1.42  & 2.41  & 2.57 \\
\rowcolor[RGB]{227,240,251}
ThinkV2V (Ours) & 5B    & 3.41  & 2.49  & 3.49  & 2.83  & 2.43  & \textbf{2.93} \\
\midrule
\multicolumn{8}{c}{\textit{Evaluation on Gemini-2.5-Pro}} \\
VACE~\cite{vace}  & 14B   & 1.49  & 1.55  & 2.07  & 1.46  & 1.26  & 1.57 \\
OmniVideo~\cite{omnivideo} & 1.3B  & 1.11  & 1.18  & 1.14  & 1.14  & 1.36  & 1.19 \\
ReCo~\cite{zhang2025region} & 1.3B  & 2.69  & 1.64  & 2.09  & 2.71  & 1.95  & 2.20 \\
InsVIE~\cite{insvie} & 2B    & 2.20  & 1.06  & 1.48  & 1.36  & 1.17  & 1.45 \\
Lucy-Edit~\cite{lucyedit} & 5B    & 2.27  & 1.57  & 3.20  & 1.75  & 2.30  & 2.22 \\
ICVE~\cite{icve}  & 13B   & 2.22  & 1.62  & 2.57  & 2.51  & 1.97  & 2.18 \\
DITTO~\cite{ditto} & 14B   & 4.01  & 1.68  & 2.03  & 1.53  & 1.41  & 2.13 \\
OpenVE-Edit~\cite{he2025openve} & 5B    & 3.16  & 2.36  & 2.98  & 1.85  & 2.15  & 2.50 \\
\rowcolor[RGB]{227,240,251}
ThinkV2V (Ours) & 5B    & 3.81  & 2.12  & 3.48  & 2.39  & 1.85  & \textbf{2.72} \\
\bottomrule
\end{tabular}
    }
    \label{tab:quant_results}
\end{table*}

\subsection{Setup}

\myparagraph{Implementation Details.}
We use Qwen3-VL-Thinking-8B~\cite{bai2025qwen3} and Wan-2.1-5B~\cite{wan2025wan} as the MLLM and DiT backbone for our ThinkV2V, respectively.
The DiT backbone is initialized from Lucy-Edit~\cite{lucyedit} to avoid re-establishing basic editing ability from scratch.
The token length of the learnable queries is set to 512 for extracting fixed-length conditioning features from the MLLM outputs.
To stabilize the early training stage, we zero-initialize the weights of the connector's final layer.
Our training is conducted on a 32-GPU cluster, following three stages.
In Stage 1, we train at 480p for 2 epochs with a learning rate of $1e^{-5}$ on OpenVE-HQ-1M.
In Stage 2, we train at 720p for 0.5 epochs with a learning rate of $1e^{-6}$ on OpenVE-HQ-1M.
In Stage 3, we continue training at 720p for 1.5 epochs with a learning rate of $1e^{-6}$ on ThinkV2V-150K.

\myparagraph{Compared Methods.}
We compare ThinkV2V with representative state-of-the-art instruction-guided video editing methods, including VACE~\cite{vace}, OmniVideo~\cite{omnivideo}, ReCo~\cite{zhang2025region}, InsVIE~\cite{insvie}, Lucy-Edit~\cite{lucyedit}, ICVE~\cite{icve}, DITTO~\cite{ditto}, and OpenVE-Edit~\cite{he2025openve}.
All methods are consistently evaluated with inference conducted on a single GPU with 80GB of VRAM.
Aligned with OpenVE-Bench~\cite{he2025openve}, for each baseline, we adopt the best resolution and frame-count configuration that can run without out-of-memory errors, so as to reflect its practical performance as fairly as possible.

\subsection{Quantitative comparison}

\begin{figure*}[t]
    \centering
    
    \includegraphics[width=1\linewidth]{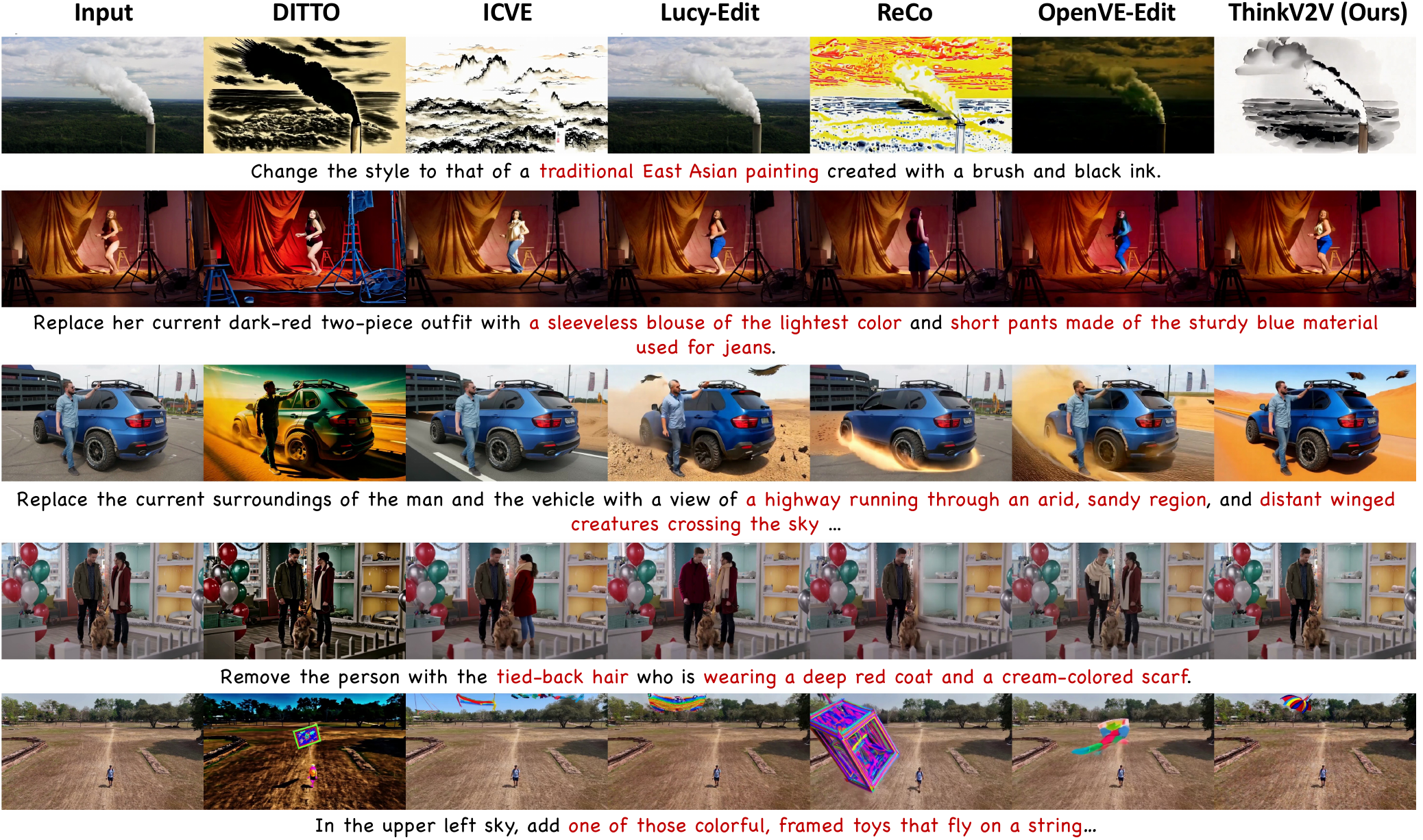}
    
    \caption
    { 
        \textbf{Qualitative comparison.}
       We provide qualitative results of our method and several competitive methods on examples from ThinkV2V-Bench across diverse reasoning-oriented editing cases, demonstrating the superior ability of ThinkV2V to produce more intention-consistent edits.
       \textit{Zoom in for better view.}
    }
    
    \label{fig:qual_results}
    
\end{figure*}

The quantitative results are reported in \cref{tab:quant_results}. We first focus on ThinkV2V-Bench, where the evaluation is based on two judge models, Seed-1.6-VL~\cite{seed2025introduction} and Gemini-2.5-Pro~\cite{comanici2025gemini}, following the OpenVE-Bench evaluation protocol~\cite{he2025openve}, and covers diverse editing tasks of Global Style, Background Change, Local Change, Local Remove, and Local Add. Several baselines exhibit competitive but localized advantages: DITTO stands out on Global Style, ReCo is particularly strong on Local Remove, and OpenVE-Edit remains competitive on Background Change. These gains, however, are mostly concentrated on specific edit types rather than being maintained across the full set of complex editing scenarios. In contrast, ThinkV2V achieves the best overall score under both evaluation settings on ThinkV2V-Bench, while remaining consistently competitive across multiple key tasks, which indicates a more extensive capability profile for complex instruction-guided video editing. Notably, despite using only a 5B-scale DiT backbone, ThinkV2V still surpasses several larger baselines through a more effective architecture design together with our training-and-inference recipe. This improvement is consistent with our central claim that explicit thinking helps bridge complex language understanding with executable visual editing.

We further report results on OpenVE-Bench \cite{he2025openve} to evaluate whether the improvement transfers to a more standard video editing benchmark. ThinkV2V continues to achieve the best overall score under both judge models on OpenVE-Bench, consistently outperforming prior methods by a clear margin while remaining especially strong on ``Local Change'' and ``Local Remove''. These results further suggest that the benefit of reasoning-driven editing is not limited to the challenging cases emphasized by ThinkV2V-Bench, but also generalizes well to more standard instruction-guided video editing scenarios.

\subsection{Qualitative comparison}

Beyond the quantitative results, we further inspect qualitative cases to understand how methods behave under complex editing instructions.
As shown in \cref{fig:qual_results}, the selected examples cover diverse edit types, including ``Global Style'', ``Local Change'', ``Background Change'',  ``Local Remove'', and ``Local Add''.
Across these cases, existing methods often misinterpret the underlying intent of the instruction, leading to incomplete style transformation, incorrect local modification, misplaced additions, or removal of the wrong target.
For example, in the ``Global Style'' case, some baselines either overfit to superficial color changes or fail to convey the intended brush-and-ink aesthetic, while in the ``Local Change'' and ``Background Change'' cases they tend to only partially satisfy the instruction or introduce visually implausible modifications.
The failure is even clearer in the ``Local Remove'' and ``Local Add'' cases, where several methods either disturb surrounding content unnecessarily or place the edited object at an incorrect location despite matching part of the textual description.
By contrast, ThinkV2V more reliably turns complex language into concrete editing actions, yielding edits that are better aligned with the full instruction while preserving unedited visual content and temporal consistency in generated videos.

\begin{table*}[t]
    \centering
    \caption{
     \textbf{Quantitative ablation studies.}
     We conduct ablations based on Seed-1.6-VL~\cite{seed2025introduction} evaluation from three key aspects, including model architecture, Progressive Curriculum Training, and Inference-Time Thinking Scaling. These results further verify the effectiveness of the proposed techniques. 
    }

    \setlength{\tabcolsep}{2.5mm}
    \renewcommand{\arraystretch}{1.2}
    
    \resizebox{\linewidth}{!}
    {
\newcommand{\methodcolwidth}{4.9cm}
\newcommand{\scorecolwidth}{2.3cm}
\newcommand{\overallcolwidth}{1.3cm}
\newcommand{\headerrowheight}{2.8ex}
\begin{tabular}{m{\methodcolwidth}|*{5}{>{\centering\arraybackslash}m{\scorecolwidth}}|>{\centering\arraybackslash}m{\overallcolwidth}}
\toprule
\rule{0pt}{\headerrowheight}Method & Global Style $\uparrow$ & Bg. Change $\uparrow$ & Local Change $\uparrow$ & Local Remove $\uparrow$ & Local Add $\uparrow$ & Overall $\uparrow$ \\
\midrule
\multicolumn{7}{c}{\textit{Ablation on Model Architecture}} \\
\midrule
w/o Thinking & 2.89  & 2.71  & 3.05  & 1.46  & 2.42  & 2.51 \\
w/ Thinking + All Features & 2.95 & 2.68 & 3.09 & 1.94 & 2.46 & 2.63 \\
\rowcolor[RGB]{227,240,251}
w/ Thinking + Answer Features  & 2.93  & 2.63  & 3.02  & 2.34  & 2.47  & \textbf{2.68} \\
\midrule
\multicolumn{7}{c}{\textit{Ablation on Progressive Curriculum Training}} \\
\midrule
Simple Only & 2.76  & 2.66  & 3.27  & 1.22  & 2.63  & 2.52 \\
Complex Only & 2.17  & 2.46  & 2.67  & 1.04  & 2.02  & 2.10 \\
Complex-to-Simple & 2.48  & 2.65  & 3.13  & 1.32  & 2.38  & 2.41 \\
Mixed Simple+Complex & 2.76  & 2.42  & 2.93  & 1.44  & 2.67  & 2.47 \\
\rowcolor[RGB]{227,240,251}
Simple-to-Complex  & 2.93  & 2.63  & 3.02  & 2.34  & 2.47  & \textbf{2.68} \\
\midrule
\multicolumn{7}{c}{\textit{Ablation on Inference-Time Thinking Scaling}} \\
\midrule
w/o Scaling & 2.93  & 2.63  & 3.02  & 2.34  & 2.47  & 2.68 \\
Serial Refinement & 2.99  & 2.67  & 3.11  & 2.18  & 2.40  & 2.67 \\
\rowcolor[RGB]{227,240,251}
Serial Refinement + Selection  & 3.13  & 2.67  & 3.07  & 2.30  & 2.42  & \textbf{2.72} \\
\bottomrule
\end{tabular}
    }
    \label{tab:ablation}
\end{table*}

\subsection{Ablation Studies and Analysis}

We conduct ablations from three aspects, including model architecture, Progressive Curriculum Training, and Inference-Time Thinking Scaling.
Unless otherwise specified, we do not enable Inference-Time Thinking Scaling in the first two groups of ablations, so that the effects of architecture and training strategy can be isolated more clearly.

\begin{figure}[t]
    \centering
    \includegraphics[width=\linewidth]{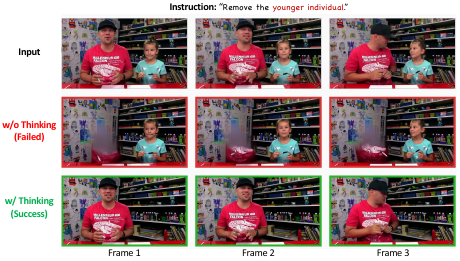}
    \caption{
        \textbf{Qualitative ablation on explicit thinking.}
        Enabling explicit thinking leads to more accurate edits, while disabling it often results in misaligned modifications under reasoning-intensive instructions.
    }
    \label{fig:qual_ablation}
\end{figure}

\myparagraph{Effect of Model Architecture.}
We first study whether explicit thinking and different choices of MLLM features affect video editing performance.
Here ``w/o Thinking'' denotes treating the MLLM as a static semantic encoder, similar to previous methods.
Thanks to training on our constructed complex data, this design attains a non-trivial overall score of 2.51.
Nevertheless, as shown in \cref{tab:ablation}, enabling explicit thinking improves the overall score from 2.51 to 2.63 when comparing ``w/o Thinking'' with ``w/ Thinking + All Features''.
Furthermore, replacing ``All Features'' with ``Answer Features'' improves the overall score to 2.68.
This suggests that using the hidden states corresponding to the final answer (\textit{e.g.}, refined prompts) is more effective than using all generated tokens, likely because it reduces redundancy from the thinking process and yields a more execution-oriented condition.
The qualitative comparison in \cref{fig:qual_ablation} further shows that enabling explicit thinking leads to more intention-consistent edits, while disabling it often results in misaligned modifications under reasoning-intensive instructions.

\myparagraph{Effect of Progressive Curriculum Training.}
We further analyze whether progressive training is more effective than alternative training schedules.
As summarized in \cref{tab:ablation}, we compare five settings: ``Simple Only'', ``Complex Only'', ``Complex-to-Simple'', ``Mixed Simple+Complex'', and ``Simple-to-Complex''.
The first two use only one data type, while the latter two either reverse the order or mix all data without staging.
Quantitatively, ``Complex Only'', ``Complex-to-Simple'', ``Mixed Simple+Complex'', and ``Simple Only'' obtain overall scores of 2.10, 2.41, 2.47, and 2.52, respectively, whereas ``Simple-to-Complex'' reaches 2.68.
This suggests that reasoning-intensive editing is best learned through a structured simple-to-complex transition, rather than from either simple or complex instructions alone.

\myparagraph{Effect of Inference-Time Thinking Scaling.}
Finally, we study whether Inference-Time Thinking Scaling can further improve performance, as summarized in \cref{tab:ablation}.
Compared with the base model without scaling, using only ``Serial Refinement'' does not improve the overall result and instead yields 2.67, likely because iterative refinement can enrich the instruction but may also amplify local ambiguities without selection.
When we add the ``Selection'' step after serial refinement, the overall score further increases to 2.72, which is the best overall result in this group.
This indicates that, in practice, iterative reasoning alone is insufficient, while the combination of iterative refinement and candidate selection is more effective at correcting potential reasoning drift and improving final editing quality.

\section{Conclusion}

In this paper, we study how to bridge implicit language intent and executable editing behaviours in instruction-guided video editing, and find that success depends not only on basic editing capabilities but also on explicit thinking.
We propose \textbf{ThinkV2V}, a reasoning-driven framework for instruction-guided video editing, and comprehensively explore three perspectives: \textit{model architecture}, \textit{training and inference recipe}, as well as \textit{dataset and benchmark construction}.
Experimental results show that ThinkV2V outperforms existing methods on reasoning-intensive editing scenarios while remaining competitive on standard video editing tasks.
Taken together, these results suggest that explicitly incorporating reasoning is a promising direction for more directly improving the overall real-world usability of instruction-guided video editing.
In the future, we would like to explore more efficient thinking-involved frameworks and further empower integrated visual content creation systems with stronger instruction understanding, planning and execution capabilities.

\clearpage
\bibliographystyle{IEEEtran}
\bibliography{main}

@String(ICML  = {Int. Conf. Mach. Learn.})

@String(AAAI  = {AAAI})

@String(ICML  = {ICML})

@article{imgedit,
  title={Imgedit: A unified image editing dataset and benchmark},
  author={Ye, Yang and He, Xianyi and Li, Zongjian and Lin, Bin and Yuan, Shenghai and Yan, Zhiyuan and Hou, Bohan and Yuan, Li},
  journal={arXiv preprint arXiv:2505.20275},
  year={2025}
}

@article{x2edit,
  title={X2Edit: Revisiting Arbitrary-Instruction Image Editing through Self-Constructed Data and Task-Aware Representation Learning},
  author={Ma, Jian and Zhu, Xujie and Pan, Zihao and Peng, Qirong and Guo, Xu and Chen, Chen and Lu, Haonan},
  journal={arXiv preprint arXiv:2508.07607},
  year={2025}
}

@article{zhang2025region,
  title={Region-Constraint In-Context Generation for Instructional Video Editing},
  author={Zhang, Zhongwei and Long, Fuchen and Li, Wei and Qiu, Zhaofan and Liu, Wu and Yao, Ting and Mei, Tao},
  journal={arXiv preprint arXiv:2512.17650},
  year={2025}
}

@article{insvie,
  title={InsViE-1M: Effective Instruction-based Video Editing with Elaborate Dataset Construction},
  author={Wu, Yuhui and Chen, Liyi and Li, Ruibin and Wang, Shihao and Xie, Chenxi and Zhang, Lei},
  journal={arXiv preprint arXiv:2503.20287},
  year={2025}
}

@article{comanici2025gemini,
  title={Gemini 2.5: Pushing the frontier with advanced reasoning, multimodality, long context, and next generation agentic capabilities},
  author={Comanici, Gheorghe and Bieber, Eric and Schaekermann, Mike and Pasupat, Ice and Sachdeva, Noveen and Dhillon, Inderjit and Blistein, Marcel and Ram, Ori and Zhang, Dan and Rosen, Evan and others},
  journal={arXiv preprint arXiv:2507.06261},
  year={2025}
}

@misc{geminiteam2026gemini31pro,
  title={{Gemini 3.1 Pro: A smarter model for your most complex tasks}},
  author={{The Gemini Team}},
  year={2026},
  month={feb},
  url={https://blog.google/innovation-and-ai/models-and-research/gemini-models/gemini-3-1-pro/},
  note={Google Blog}
}

@article{seed2025introduction,
  title={Introduction to techniques used in seed1. 6},
  author={Seed, ByteDance},
  journal={Online. Accessed},
  pages={12--21},
  year={2025}
}

@article{step1xedit,
  title={Step1x-edit: A practical framework for general image editing},
  author={Liu, Shiyu and Han, Yucheng and Xing, Peng and Yin, Fukun and Wang, Rui and Cheng, Wei and Liao, Jiaqi and Wang, Yingming and Fu, Honghao and Han, Chunrui and others},
  journal={arXiv preprint arXiv:2504.17761},
  year={2025}
}

@article{omnigen2,
  title={OmniGen2: Exploration to Advanced Multimodal Generation},
  author={Wu, Chenyuan and Zheng, Pengfei and Yan, Ruiran and Xiao, Shitao and Luo, Xin and Wang, Yueze and Li, Wanli and Jiang, Xiyan and Liu, Yexin and Zhou, Junjie and others},
  journal={arXiv preprint arXiv:2506.18871},
  year={2025}
}

@article{qwenimage,
  title={Qwen-image technical report},
  author={Wu, Chenfei and Li, Jiahao and Zhou, Jingren and Lin, Junyang and Gao, Kaiyuan and Yan, Kun and Yin, Sheng-ming and Bai, Shuai and Xu, Xiao and Chen, Yilei and others},
  journal={arXiv preprint arXiv:2508.02324},
  year={2025}
}

@article{omnivideo,
  title={Omni-video: Democratizing unified video understanding and generation},
  author={Tan, Zhiyu and Yang, Hao and Qin, Luozheng and Gong, Jia and Yang, Mengping and Li, Hao},
  journal={arXiv preprint arXiv:2507.06119},
  year={2025}
}

@article{univideo,
  title={UniVideo: Unified Understanding, Generation, and Editing for Videos},
  author={Wei, Cong and Liu, Quande and Ye, Zixuan and Wang, Qiulin and Wang, Xintao and Wan, Pengfei and Gai, Kun and Chen, Wenhu},
  journal={arXiv preprint arXiv:2510.08377},
  year={2025}
}

@article{instructx,
  title={InstructX: Towards Unified Visual Editing with MLLM Guidance},
  author={Mou, Chong and Sun, Qichao and Wu, Yanze and Zhang, Pengze and Li, Xinghui and Ye, Fulong and Zhao, Songtao and He, Qian},
  journal={arXiv preprint arXiv:2510.08485},
  year={2025}
}

@article{lucyedit,
  title   = {Lucy Edit: Open-Weight Text-Guided Video Editing},
  author  = {DecartAI Team},
  year    = {2025},
  url     = {https://d2drjpuinn46lb.cloudfront.net/Lucy_Edit__High_Fidelity_Text_Guided_Video_Editing.pdf}
 }

@misc{icve,
      title={In-Context Learning with Unpaired Clips for Instruction-based Video Editing}, 
      author={Xinyao Liao and Xianfang Zeng and Ziye Song and Zhoujie Fu and Gang Yu and Guosheng Lin},
      year={2025},
      eprint={2510.14648},
      archivePrefix={arXiv},
      primaryClass={cs.CV},
      url={https://arxiv.org/abs/2510.14648}, 
}

@article{kontext,
  title={FLUX. 1 Kontext: Flow Matching for In-Context Image Generation and Editing in Latent Space},
  author={Labs, Black Forest and Batifol, Stephen and Blattmann, Andreas and Boesel, Frederic and Consul, Saksham and Diagne, Cyril and Dockhorn, Tim and English, Jack and English, Zion and Esser, Patrick and others},
  journal={arXiv preprint arXiv:2506.15742},
  year={2025}
}

@inproceedings{huang2024vbench,
  title={Vbench: Comprehensive benchmark suite for video generative models},
  author={Huang, Ziqi and He, Yinan and Yu, Jiashuo and Zhang, Fan and Si, Chenyang and Jiang, Yuming and Zhang, Yuanhan and Wu, Tianxing and Jin, Qingyang and Chanpaisit, Nattapol and others},
  booktitle={Proceedings of the IEEE/CVF Conference on Computer Vision and Pattern Recognition},
  pages={21807--21818},
  year={2024}
}

@inproceedings{radford2021learning,
  title={Learning transferable visual models from natural language supervision},
  author={Radford, Alec and Kim, Jong Wook and Hallacy, Chris and Ramesh, Aditya and Goh, Gabriel and Agarwal, Sandhini and Sastry, Girish and Askell, Amanda and Mishkin, Pamela and Clark, Jack and others},
  booktitle={International conference on machine learning},
  pages={8748--8763},
  year={2021},
  organization={PmLR}
}

@misc{ditto,
      title={Scaling Instruction-Based Video Editing with a High-Quality Synthetic Dataset}, 
      author={Qingyan Bai and Qiuyu Wang and Hao Ouyang and Yue Yu and Hanlin Wang and Wen Wang and Ka Leong Cheng and Shuailei Ma and Yanhong Zeng and Zichen Liu and Yinghao Xu and Yujun Shen and Qifeng Chen},
      year={2025},
      archivePrefix={arXiv},
      primaryClass={cs.CV},
      url={https://arxiv.org/abs/2510.15742}, 
}

@article{vace,
  title={Vace: All-in-one video creation and editing},
  author={Jiang, Zeyinzi and Han, Zhen and Mao, Chaojie and Zhang, Jingfeng and Pan, Yulin and Liu, Yu},
  journal={arXiv preprint arXiv:2503.07598},
  year={2025}
}

@article{kiwiedit,
  title={Kiwi-Edit: Versatile Video Editing via Instruction and Reference Guidance},
  author={Lin, Yiqi and Liang, Guoqiang and Zeng, Ziyun and Bai, Zechen and Chen, Yanzhe and Shou, Mike Zheng},
  journal={arXiv preprint arXiv:2603.02175},
  year={2026}
}

@article{sama,
  title={SAMA: Factorized Semantic Anchoring and Motion Alignment for Instruction-Guided Video Editing},
  author={Zhang, Xinyao and Dong, Wenkai and Song, Yuxin and Fang, Bo and Zhang, Qi and Wang, Jing and Chen, Fan and Zhang, Hui and Feng, Haocheng and Lu, Yu and others},
  journal={arXiv preprint arXiv:2603.19228},
  year={2026}
}

@article{revise,
  title={ReViSE: Towards Reason-Informed Video Editing in Unified Models with Self-Reflective Learning},
  author={Liu, Xinyu and Yuan, Hangjie and Wei, Yujie and Xing, Jiazheng and Han, Yujin and Pan, Jiahao and Ma, Yanbiao and Chan, Chi-Min and Zhao, Kang and Zhang, Shiwei and others},
  journal={arXiv preprint arXiv:2512.09924},
  year={2025}
}

@article{openve,
  title={OpenVE-3M: A Large-Scale High-Quality Dataset for Instruction-Guided Video Editing},
  author={He, Haoyang and Wang, Jie and Zhang, Jiangning and Xue, Zhucun and Bu, Xingyuan and Yang, Qiangpeng and Wen, Shilei and Xie, Lei},
  journal={arXiv preprint arXiv:2512.07826},
  year={2025}
}

@inproceedings{llavacot,
  title={Llava-cot: Let vision language models reason step-by-step},
  author={Xu, Guowei and Jin, Peng and Wu, Ziang and Li, Hao and Song, Yibing and Sun, Lichao and Yuan, Li},
  booktitle={Proceedings of the IEEE/CVF International Conference on Computer Vision},
  pages={2087--2098},
  year={2025}
}

@article{bai2025qwen3,
  title={Qwen3-vl technical report},
  author={Bai, Shuai and Cai, Yuxuan and Chen, Ruizhe and Chen, Keqin and Chen, Xionghui and Cheng, Zesen and Deng, Lianghao and Ding, Wei and Gao, Chang and Ge, Chunjiang and others},
  journal={arXiv preprint arXiv:2511.21631},
  year={2025}
}

@article{wan2025wan,
  title={Wan: Open and advanced large-scale video generative models},
  author={Wan, Team and Wang, Ang and Ai, Baole and Wen, Bin and Mao, Chaojie and Xie, Chen-Wei and Chen, Di and Yu, Feiwu and Zhao, Haiming and Yang, Jianxiao and others},
  journal={arXiv preprint arXiv:2503.20314},
  year={2025}
}

@article{ye2025blink,
  title={BLINK-Twice: You see, but do you observe? A Reasoning Benchmark on Visual Perception},
  author={Ye, Junyan and Jiang, Dongzhi and He, Jun and Zhou, Baichuan and Huang, Zilong and Yan, Zhiyuan and Li, Hongsheng and He, Conghui and Li, Weijia},
  journal={arXiv preprint arXiv:2510.09361},
  year={2025}
}

@article{huang2025vchain,
  title={Vchain: Chain-of-visual-thought for reasoning in video generation},
  author={Huang, Ziqi and Yu, Ning and Chen, Gordon and Qiu, Haonan and Debevec, Paul and Liu, Ziwei},
  journal={arXiv preprint arXiv:2510.05094},
  year={2025}
}

@article{hao2025enhancing,
  title={Enhancing physical plausibility in video generation by reasoning the implausibility},
  author={Hao, Yutong and Chen, Chen and Mian, Ajmal Saeed and Xu, Chang and Liu, Daochang},
  journal={arXiv preprint arXiv:2509.24702},
  year={2025}
}

@inproceedings{yang2024mastering,
  title={Mastering Text-to-Image Diffusion: Recaptioning, Planning, and Generating with Multimodal LLMs.},
  author={Yang, Ling and Yu, Zhaochen and Meng, Chenlin and Xu, Minkai and Ermon, Stefano and Cui, Bin},
  booktitle={Icml},
  volume={3},
  number={6},
  pages={7},
  year={2024}
}

@article{he2025openve,
  title={OpenVE-3M: A Large-Scale High-Quality Dataset for Instruction-Guided Video Editing},
  author={He, Haoyang and Wang, Jie and Zhang, Jiangning and Xue, Zhucun and Bu, Xingyuan and Yang, Qiangpeng and Wen, Shilei and Xie, Lei},
  journal={arXiv preprint arXiv:2512.07826},
  year={2025}
}

@article{zhou2024magictailor,
  title={Magictailor: Component-controllable personalization in text-to-image diffusion models},
  author={Zhou, Donghao and Huang, Jiancheng and Bai, Jinbin and Wang, Jiaze and Chen, Hao and Chen, Guangyong and Hu, Xiaowei and Heng, Pheng-Ann},
  journal={arXiv preprint arXiv:2410.13370},
  year={2024}
}

@article{liu2026hifi,
  title={HiFi-Inpaint: Towards High-Fidelity Reference-Based Inpainting for Generating Detail-Preserving Human-Product Images},
  author={Liu, Yichen and Zhou, Donghao and Wang, Jie and Gao, Xin and Liu, Guisheng and Li, Jiatong and Zhang, Quanwei and Lyu, Qiang and Guo, Lanqing and Wen, Shilei and others},
  journal={arXiv preprint arXiv:2603.02210},
  year={2026}
}

@inproceedings{zhou2026identitystory,
  title={IdentityStory: Taming Your Identity-Preserving Generator for Human-Centric Story Generation},
  author={Zhou, Donghao and Lin, Jingyu and Shen, Guibao and Liu, Quande and Gao, Jialin and Liu, Lihao and Du, Lan and Chen, Cunjian and Fu, Chi-Wing and Hu, Xiaowei and others},
  booktitle={Proceedings of the AAAI Conference on Artificial Intelligence},
  volume={40},
  number={16},
  pages={13593--13601},
  year={2026}
}

@article{zhou2026omnishow,
  title={Omnishow: Unifying multimodal conditions for human-object interaction video generation},
  author={Zhou, Donghao and Liu, Guisheng and Yang, Hao and Li, Jiatong and Lin, Jingyu and Huang, Xiaohu and Liu, Yichen and Gao, Xin and Chen, Cunjian and Wen, Shilei and others},
  journal={arXiv preprint arXiv:2604.11804},
  year={2026}
}

@article{song2025scenedecorator,
  title={SceneDecorator: Towards Scene-Oriented Story Generation with Scene Planning and Scene Consistency},
  author={Song, Quanjian and Zhou, Donghao and Lin, Jingyu and Shen, Fei and Wang, Jiaze and Hu, Xiaowei and Chen, Cunjian and Heng, Pheng-Ann},
  journal={arXiv preprint arXiv:2510.22994},
  year={2025}
}

@article{chen2025empirical,
  title={An empirical study of gpt-4o image generation capabilities},
  author={Chen, Sixiang and Bai, Jinbin and Zhao, Zhuoran and Ye, Tian and Shi, Qingyu and Zhou, Donghao and Chai, Wenhao and Lin, Xin and Wu, Jianzong and Tang, Chao and others},
  journal={arXiv preprint arXiv:2504.05979},
  year={2025}
}

\clearpage

\appendices

\noindent{\normalfont\Large\sffamily\bfseries\scshape Appendices}
\vspace{0.4em}

\noindent The appendices present the following sections to complement the main manuscript:
\begin{itemize}[leftmargin=2em, itemsep=0.15em, topsep=-0.2em]
    \item \cref{sec:data_benchmark_details} details the construction of ThinkV2V-150K and ThinkV2V-Bench.
    \item \cref{sec:more_qualitative_results} provides additional qualitative comparisons on ThinkV2V-Bench and more challenging cases for ThinkV2V.
    \item \cref{sec:limitations_future_work} discusses the current limitations of ThinkV2V and outlines several promising future directions.
    \item \cref{sec:impact_statement} presents the broader societal impacts of ThinkV2V.
\end{itemize}

\section{Details of Dataset and Benchmark Construction}
\label{sec:data_benchmark_details}

\subsection{ThinkV2V-150K}
\label{sec:appendix_dataset}

\myparagraph{Motivation.}
A central goal of this work is to systematically study reasoning-driven video editing rather than editing based only on direct target matching. In realistic user requests, the desired editing target is often not explicitly specified by its appearance alone. Instead, the instruction may refer to an object, region, or transformation through contextual cues such as spatial relations, object interactions, scene dynamics, function, or causal role in the scene. As a result, building training data for this setting requires more than collecting visually aligned source-target pairs: it also requires constructing high-quality video pairs and selecting samples whose editing intent can be naturally rewritten into indirect, reasoning-oriented language.

\myparagraph{Category Filtering.}
We construct ThinkV2V-150K from OpenVE-3M. Among the original OpenVE categories, we retain only five: ``Global Style'', ``Background Change'', ``Local Remove'', ``Local Add'', and ``Local Change''. We choose these categories because they cover a broad range of useful editing operations while keeping the transformation grounded in the source video. They span both scene-level and object-level edits, making them suitable for studying whether a model can connect high-level language intent to concrete editing behavior.
We exclude the remaining three categories, namely ``Camera Edit'', ``Subtitle Edit'', and ``Creative Edit''. These categories are less aligned with our objective of evaluating reasoning-centric editing. ``Camera Edit'' mainly concerns viewpoint or camera control, ``Subtitle Edit'' is tied to text insertion and rendering, and ``Creative Edit'' often involves highly open-ended transformations whose outputs are comparatively less grounded in reasoning about scene entities, relations, and causal structure. Since our goal is to focus on edits that require identifying and modifying targets through implicit intent and contextual reasoning, these categories are excluded from both training data construction and benchmark design.

\myparagraph{Quality Filtering.}
We first perform video quality filtering on OpenVE-3M to construct OpenVE-HQ-1M. Since large-scale video editing data can contain noisy, visually implausible, or temporally unstable samples, this stage aims to obtain a cleaner source pool before reasoning-oriented instruction synthesis. Specifically, we rescore each sample using Gemini-2.5-Flash~\cite{comanici2025gemini} to estimate whether the edited result follows the instruction and whether the edit is visually plausible. Based on this score, we retain the top 50\% samples as an initial high-quality candidate pool.
To account for temporal consistency, we compute two automatic video-level quality signals for each edited video: the average inter-frame CLIP similarity (CLIP-F)~\cite{radford2021learning} and Temporal Flickering (TF)~\cite{huang2024vbench}. CLIP-F measures semantic consistency across frames, while TF quantifies frame-to-frame flickering artifacts introduced by the edit. These metrics complement semantic scoring by capturing whether the edited video remains coherent over time.
We then combine Gemini-based scoring with CLIP-F and TF to remove samples with weak instruction alignment, low inter-frame semantic consistency, or strong temporal flickering. This multi-criteria filtering yields OpenVE-HQ-1M, a high-quality subset of 1M video pairs that serves as the source pool for subsequent suitability filtering and instruction rewriting.
This step is important because reasoning-oriented instruction following is difficult to learn from noisy supervision. If the visual edit is poorly aligned with the instruction, visually implausible, or temporally unstable, the model may fail to distinguish whether errors arise from weak language understanding or low-quality training pairs. By filtering OpenVE-3M with both semantic and temporal quality signals, OpenVE-HQ-1M provides a cleaner foundation for constructing ThinkV2V-150K.

\myparagraph{Suitability Filtering.}
From OpenVE-HQ-1M, we further select samples that are suitable for reasoning-instruction synthesis. The exact filtering prompt used in this stage is provided in \cref{prompt:stage1}. For each candidate sample, Gemini-2.5-Flash~\cite{comanici2025gemini} is given the first frame of the original video, the first frame of the edited video, and the original editing instruction, and is asked to return a JSON object containing both suitability and quality judgments.
The suitability judgment is used to determine whether the original editing intent can be naturally rewritten into a more complex, indirect, reasoning-based description. In particular, we favor samples whose target object, style, action, or transformation can be described through features, functions, attributes, contextual relations, or associated knowledge rather than by directly naming the target. This removes samples whose instructions are too literal, too ambiguous, or lack sufficient semantic room for reasoning-oriented rewriting.
The quality judgment in \cref{prompt:stage1} serves as an additional sample-level sanity check during this selection stage. Although OpenVE-HQ-1M has already been filtered by semantic and temporal quality signals, this prompt-level judgment helps further remove cases where the edit is difficult to verify from the first-frame visual context or where the instruction--video correspondence is not sufficiently clear for reliable instruction rewriting. After this suitability filtering, we retain 20K to 40K samples per category, resulting in 150K video pairs in total.

\myparagraph{Instruction Rewriting.}
We then use Gemini-2.5-Pro~\cite{comanici2025gemini} to rewrite the original direct instructions into more complex reasoning-oriented ones. The exact rewriting prompt is shown in \cref{prompt:stage2}. As specified in \cref{prompt:stage2}, the rewriting process is conditioned on the original instruction together with the visual context from the first frames of the source and edited videos.
The prompt explicitly asks the model to: (1) remove generic quality-control phrases such as ``ensure consistency'' or ``keep the background unchanged,'' (2) replace direct naming with indirect references based on features, functions, attributes, or associated knowledge, (3) keep the rewritten instruction natural and user-like rather than definitional, and (4) handle all editing types in a consistent manner. These constraints are designed to preserve the intended edit outcome while shifting the linguistic form from direct commands to more realistic, inference-heavy user requests.
Compared with simple appearance-based prompts, the resulting instructions place greater emphasis on implicit goals, contextual references, relational cues, and causal understanding. This increases the gap between surface-level wording and the actual executable editing target, thereby encouraging the model to reason over scene content rather than relying only on literal phrase matching. The resulting ThinkV2V-150K dataset therefore exposes models to instruction distributions that are closer to realistic user requests and promotes editing grounded in implicit intent and causal reasoning.

\prompt{Data Filtering}{ 
 You are a video editing data quality assessor. 
 You will be given: 
 \begin{itemize}
   \item The first frame of the original video. 
   \item The first frame of the edited video. 
   \item The editing instruction. 
 \end{itemize}
 Please answer the following two questions: 
 \begin{enumerate}
   \item Suitability: Is this editing instruction suitable for being rewritten into a more complex, 
    indirect, reasoning-based description? (i.e., the object/style/action can be described 
    by its features, function, or associated knowledge rather than its name directly.) 
    Answer Yes or No, then briefly explain in $\le$ 15 words. 
   \item Quality: Is the editing quality of this video pair high? (The edit should clearly follow 
    the instruction, look visually natural, and be temporally stable.) 
    Answer Yes or No, then briefly explain in $\le$ 15 words. 
 \end{enumerate}
 Output ONLY a valid JSON object in the following format, with no extra text: 
 
 \vspace{0.5em}
 \noindent\texttt{\{} \\
 \texttt{\ \ "suitable": true or false,} \\
 \texttt{\ \ "suitable\_reason": "...",} \\
 \texttt{\ \ "quality": true or false,} \\
 \texttt{\ \ "quality\_reason": "..."} \\
 \texttt{\}}
 }{prompt:stage1}

\subsection{ThinkV2V-Bench}
\label{sec:appendix_benchmark}

\myparagraph{Motivation.}
Existing benchmarks are useful for measuring basic editing quality, but they are less suitable for evaluating whether a model understands implicit intent in language. In many standard settings, the instruction directly names the target object, attribute, or desired transformation, which makes it possible to achieve strong performance through shallow matching between text and visible content. However, such settings do not adequately test whether the model can infer the intended editing condition from contextual or relational descriptions. To evaluate this ability more directly, we construct ThinkV2V-Bench as a benchmark tailored to reasoning-driven video editing.

\myparagraph{Category Filtering.}
For consistency with ThinkV2V-150K, we benchmark only the same five reasoning-suitable categories: ``Global Style'', ``Background Change'', ``Local Remove'', ``Local Add'', and ``Local Change''. We likewise exclude ``Camera Edit'', ``Subtitle Edit'', and ``Creative Edit''. This alignment keeps the task definition coherent across training and evaluation, and ensures that benchmark results reflect reasoning ability within the intended editing regime rather than performance on camera manipulation, text rendering, or highly subjective creative generation.

\myparagraph{Instruction Rewriting.}
We build ThinkV2V-Bench from OpenVE-Bench by rewriting the original direct instructions into reasoning-oriented ones. To keep benchmark construction aligned with the training data pipeline, we follow the same rewriting principle as \cref{prompt:stage2}, while using Gemini-3.1-Pro~\cite{geminiteam2026gemini31pro} for the benchmark rewriting stage. For each source video, edited video, and instruction triplet, the model receives the original instruction together with the visual context and rewrites it into a more indirect, inference-heavy instruction that preserves the same intended edit.
In this way, the benchmark shifts the focus away from literal target naming and toward identification through context, relation, position, action, function, or other implicit cues. As a result, success on ThinkV2V-Bench depends not only on producing visually plausible edits, but also on correctly translating complex language understanding into the precise editing condition required by the reference output.
This process yields 308 video-instruction pairs across the five selected categories. Although compact in size, ThinkV2V-Bench is deliberately targeted: its purpose is not to maximize scale, but to provide a focused testbed for validating whether a model trained for reasoning-driven editing can actually execute edits implied by indirect instructions.

 \prompt{Complex Editing Instruction Rewriting}{ 
 You are a creative video editing instruction rewriting expert. 
 Your task is to rewrite a simple, direct video editing instruction into a complex, reasoning-based instruction based on the visual context of the video's first frame. 
 \vspace{0.5em}\newline\noindent\textbf{Constraints}: 
 \begin{enumerate}
   \item \textbf{No Quality Constraints}: Do NOT add phrases like ``ensure consistency'', ``high quality'', ``realistic lighting'', ``keep the background unchanged'', etc. Focus ONLY on the content of the edit. If the input instruction contains such constraints, REMOVE them. 
   \item \textbf{Indirect Referencing}: Instead of naming the object or style directly, describe its features, functions, attributes, or associated knowledge. 
   \item \textbf{Natural Style}: The output should sound like a user describing what they want in a slightly roundabout or descriptive way, not a dictionary definition. 
  \item \textbf{Coverage}: Handle all editing types (``Global Style'', ``Local Change'', ``Local Remove'', ``Local Add'', etc.). 
 \end{enumerate}
 \vspace{0.5em}\noindent\textbf{Examples}: 
 \begin{center} 
\begingroup
\scriptsize
\setlength{\tabcolsep}{1.8pt}
\renewcommand{\arraystretch}{1.08}
\begin{tabularx}{\linewidth}{p{1.15cm} p{1.75cm} p{1.75cm} X} 
 \hline 
 \textbf{Edit Type} & \textbf{Visual Context} & \textbf{Simple Instruction} & \textbf{Complex Instruction} \\ 
 \hline 
 \textbf{Global Style} & (N/A) & Change to oil painting style. & Transform the visual into a textured canvas art style with visible brushstrokes. \\ 
 \hline 
 \textbf{Local Change} & A woman wearing a skirt. & Turn the skirt into denim. & Give the skirt the texture and appearance of the sturdy blue fabric typically used for jeans. \\ 
 \hline 
 \textbf{Local Remove} & Two people: an old man on the left, a young man on the right. & Remove the person on the left. & Remove the older individual from the scene. \\ 
 \hline 
 \textbf{Local Remove} & A person wearing a face mask. & Remove the mask. & Take off the facial covering used for protection. \\ 
 \hline 
 \textbf{Local Add} & An intersection. & Add a traffic light. & Insert a three-color signaling device used for traffic control. \\ 
 \hline 
 \textbf{Local Add} & A plate on a table. & Add sushi to the plate. & Place a serving of the Japanese rice dish usually wrapped in seaweed on the plate. \\ 
 \hline 
 \textbf{Local Add} & An empty table. & Add a coffee. & Place a cup of dark, energizing hot beverage on the table. \\ 
 \hline 
\end{tabularx}
\endgroup
 \end{center}
 You will receive the ``Input Instruction'' and the ``Visual Context'' (first frames of the original and edited videos). 
 Output ONLY the rewritten complex instruction. Do not output any explanation or extra text. 
 }{prompt:stage2}

\subsection{Scope and Summary}
In \cref{fig:data_bench}, we summarize the key statistics of ThinkV2V-150K, and also present a representative example of ThinkV2V-Bench.
Together, ThinkV2V-150K and ThinkV2V-Bench form a closed data-and-evaluation loop for our study. ThinkV2V-150K provides large-scale supervision that shifts the training distribution toward realistic, reasoning-oriented user requests, while ThinkV2V-Bench offers a controlled benchmark for assessing whether that training translates into better execution of implicit intent. This pairing enables systematic training and validation of our core claim: effective video editing should rely not only on visual generation quality, but also on the ability to reason over language, scene context, and causal structure.

\section{More Qualitative Results}
\label{sec:more_qualitative_results}

\begin{figure*}[t]
    \centering
    
    \includegraphics[width=1\linewidth]{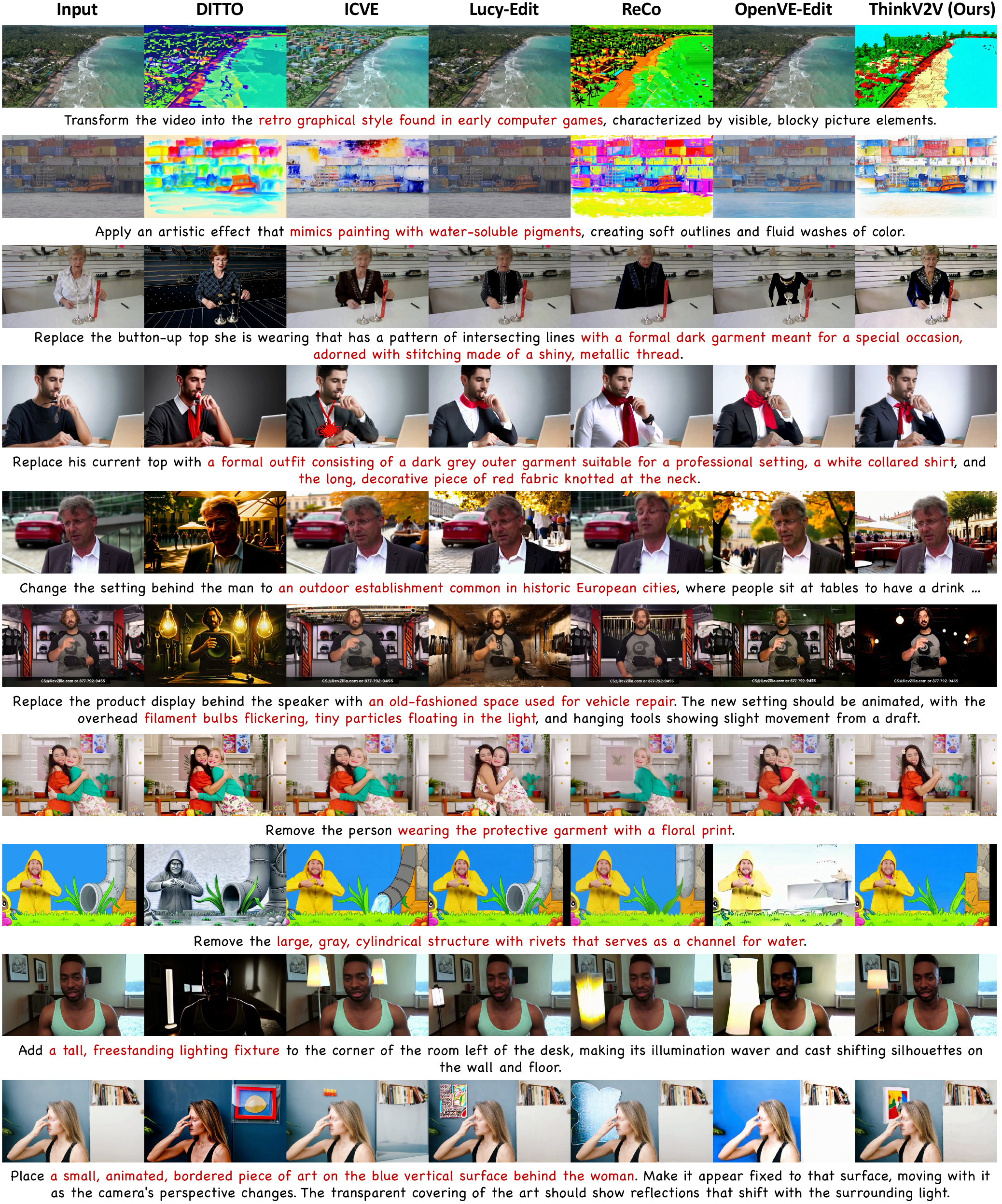}
    
    \caption
    {
        \textbf{More qualitative comparison.}
        We provide additional qualitative results of ThinkV2V and other competitive methods on more examples from ThinkV2V-Bench, demonstrating the superior ability of ThinkV2V to produce more intention-consistent edits.
        \textit{Zoom in for better view.}
    }
    
    \label{fig:qual_results_more}
    
\end{figure*}

\begin{figure*}[t]
    \centering
    
    \includegraphics[width=1\linewidth]{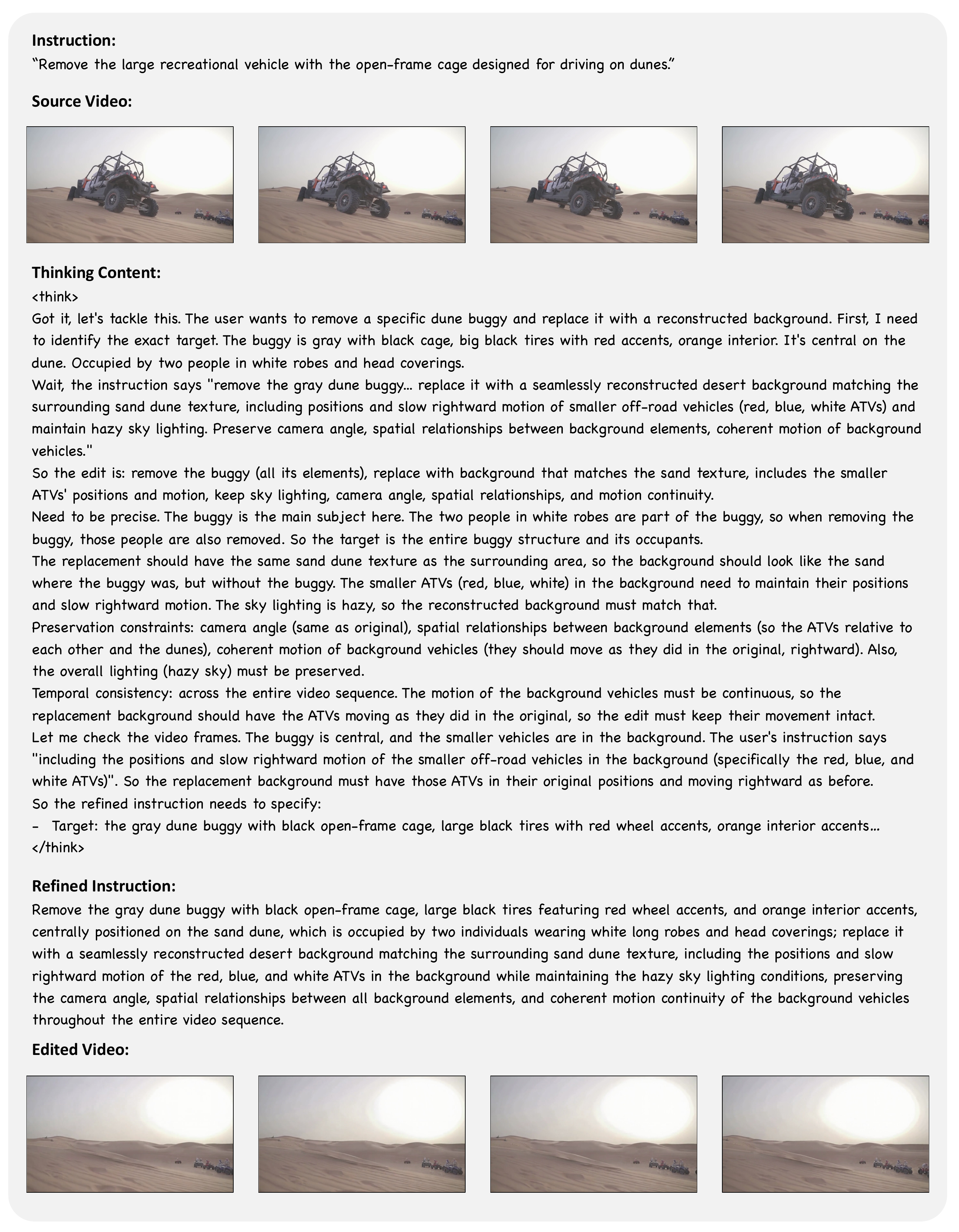}
    
    \caption
    {
        \textbf{ThinkV2V on a challenging reasoning-intensive case.}
        A detailed example showing how ThinkV2V handles a complex instruction with implicit target identification, fine-grained scene understanding, and strong consistency requirements.
        \textit{Zoom in for better view.}
    }
    
    \label{fig:chall_case_1}
    
\end{figure*}

\begin{figure*}[t]
    \centering
    
    \includegraphics[width=1\linewidth]{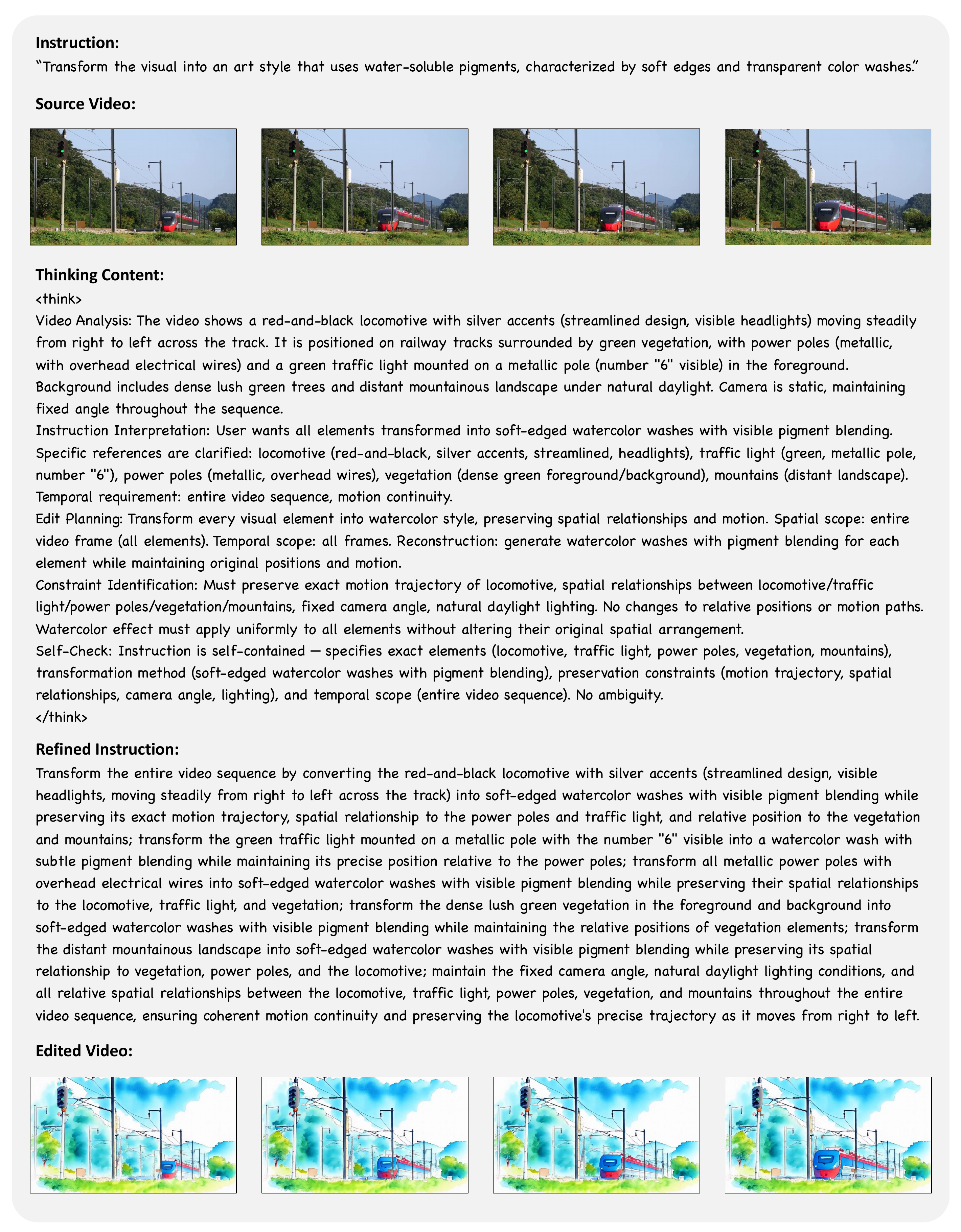}
    
    \caption
    {
        \textbf{ThinkV2V on a challenging reasoning-intensive case.}
        A detailed example showing how ThinkV2V handles a complex instruction with implicit target identification, fine-grained scene understanding, and strong consistency requirements.
        \textit{Zoom in for better view.}
    }
    
    \label{fig:chall_case_2}
    
\end{figure*}

To complement the qualitative results in the main paper, we present more video editing cases to more fully examine the performance improvement of ThinkV2V over existing methods.
In \cref{fig:qual_results_more}, we provide additional qualitative comparisons on ThinkV2V-Bench, covering representative scenarios including ``Global Style'', ``Background Change'', ``Local Remove'', ``Local Add'', and ``Local Change''.
Consistent with the observations in the main paper, existing methods still tend to edit the wrong target, drift to unintended local regions, or follow surface keywords without fully executing the underlying intent.
\cref{fig:chall_case_1} presents a challenging ``Local Remove'' case, where the instruction refers to the target dune buggy through detailed visual attributes, preserves other vehicles, and simultaneously requires seamless background reconstruction together with preservation of the positions and slow motion of other off-road vehicles.
\cref{fig:chall_case_2} presents a challenging ``Global Style'' case, where the entire video needs to be transformed into a watercolor-like style while preserving the locomotive, traffic light, power poles, vegetation, mountains, their spatial relationships, and the train's motion trajectory.
These two cases show that, through explicit thinking, ThinkV2V can not only resolve indirect references to the true editing target, but also maintain scene structure, motion continuity, and global consistency under stronger semantic and temporal constraints.
Overall, these additional qualitative results further support the advantage of ThinkV2V in reasoning-driven video editing under complex real-world instructions.

\section{Limitations and Future Work}
\label{sec:limitations_future_work}

Although ThinkV2V achieves competitive performance on instruction-guided video editing, the current study still has several limitations worth further improvement.
First, explicit thinking inevitably introduces additional computation cost and latency, which increases both training and inference overhead.
Second, although final editing performance supports the usefulness of explicit thinking, we do not independently evaluate thinking content correctness because rigorous quantitative evaluation protocols are unavailable for semantic reasoning traces.
Third, although ThinkV2V-150K and ThinkV2V-Bench are specifically designed for implicit intent understanding and causal reasoning, their current coverage is still limited for more open-domain, longer-horizon, and rarer complex editing scenarios.

Looking forward, one promising direction is to develop more efficient reasoning-involved video editing frameworks, for example by exploring latent reasoning techniques, so as to reduce the extra cost introduced by explicit thinking.
It is also important to expand the training data and benchmark to cover richer reasoning patterns, editing scenarios, and video domains.
Together, these directions may further empower integrated visual content creation systems with stronger user-provided instruction understanding, planning, and execution capabilities, bringing them closer to more general and reliable multimodal creative assistants at scale.

\section{Impact Statement}
\label{sec:impact_statement}

The proposed ThinkV2V targets instruction-guided video editing, whose broader impact includes both benefits for visual content creation and risks commonly associated with generative video models.
On the positive side, it can improve creative and educational content production by reducing manual trial and error in complex editing workflows.
At the same time, stronger video editing capabilities may be misused to create misleading or impersonation content, and such misuse may further amplify risks related to bias, privacy, and authorization.
We therefore plan to adopt a responsible release strategy with clear usage terms, restrictions on high-risk use cases, and necessary documentation of licenses and intended use to help support safer and more responsible use.

\end{document}